\documentclass[letterpaper, 10 pt, conference]{ieeeconf} 

\IEEEoverridecommandlockouts    % This command is only needed if 
\usepackage{multicol}

\usepackage{outlines}

\usepackage{url}
\usepackage{graphicx}
\usepackage{amsmath}
\usepackage{amsfonts}
\usepackage{subcaption}
\usepackage{makecell}
\usepackage{xcolor}
\usepackage{cite}
\usepackage{float}
\usepackage{hyperref}

\newcommand{\mb}[1]{\mathbf{#1}}

\begin{document}

\title{\LARGE \bf Learning Hybrid Object Kinematics for Efficient Hierarchical Planning Under Uncertainty}

% \author{
%   Ajinkya Jain \\
%   Department of Mechanical Engineering\\
%     University of Texas at Austin, USA\\
%   \texttt{ajinkya@utexas.edu} \\
%     \and
%   Scott Niekum\\
%   Department of Computer Science\\
%   University of Texas at Austin, USA\\
%   \texttt{sniekum@cs.utexas.edu} \\
% }

\author{Ajinkya Jain$^{1}$ and Scott Niekum$^{1}$%
\thanks{$^{1}$Personal Autonomous Robotics Lab (PeARL), The University of Texas at Austin. Contact: {\tt\small ajinkya@utexas.edu}}}%

\maketitle

\thispagestyle{empty}
\pagestyle{empty}

%%%%%%%%%%%%%%%%%%%%%%%%%%%%%%%%%%%%%%%%%%%%%%%%%%%%%%%%%%%%%%%%%%%%%%%%%%%%%%%%
\begin{abstract}
Sudden changes in the dynamics of robotic tasks, such as contact with an object or the latching of a door, are often viewed as inconvenient discontinuities that make manipulation difficult. However, when these transitions are well-understood, they can be leveraged to reduce uncertainty or aid manipulation---for example, wiggling a screw to determine if it is fully inserted or not. Current model-free reinforcement learning approaches require large amounts of data to learn to leverage such dynamics, scale poorly as problem complexity grows, and do not transfer well to significantly different problems. By contrast, hierarchical POMDP planning-based methods scale well via plan decomposition, work well on novel problems, and directly consider uncertainty, but often rely on precise hand-specified models and task decompositions. To combine the advantages of these opposing paradigms, we propose a new method, MICAH, which given unsegmented data of an object's motion under applied actions, (1) detects changepoints in the object motion model using action-conditional inference, (2) estimates the individual local motion models with their parameters, and (3) converts them into a hybrid automaton that is compatible with hierarchical POMDP planning. We show that model learning under MICAH is more accurate and robust to noise than prior approaches. Further, we combine MICAH with a hierarchical POMDP planner to demonstrate that the learned models are rich enough to be used for performing manipulation tasks under uncertainty that require the objects to be used in novel ways not encountered during training. 
\end{abstract}

%%%%%%%%%%%%%%%%%%%%%%%%%%%%%%%%%%%%%%%%%%%%%%%%%%%%%%%%%%%%%%%%%%%%%%%%%%%%%%%%

\section{Introduction} \label{intro}
Robots working in human environments need to perform dexterous manipulation on a wide variety of objects. Such tasks typically involve making or breaking contacts with other objects, leading to sudden discontinuities in the task dynamics. Furthermore, many objects exhibit configuration-dependent dynamics, such as a refrigerator door that stays closed magnetically. While the presence of such nonlinearities in task dynamics can make it challenging to represent good manipulation policies and models, if well-understood, these nonlinearities can also be leveraged to improve task performance and reduce uncertainty. For example, when inserting a screw into the underside of a table, if direct visual feedback is not available, indirect feedback from wiggling the screw (a semi-rigid connection between the screw and the table) can be leveraged to ascertain whether the screw is inserted or not.  In other words, the sensed change in dynamics (from free-body motion to rigid contact) serves as a landmark, partially informing the robot about the state of the system and reducing uncertainty. Such dynamics can be naturally represented as hybrid dynamics models or \textit{hybrid automata} \cite{lygeros2012hybrid}, in which a discrete state represents which continuous dynamics model is active at any given time.

Current model-free reinforcement learning approaches \cite{pmlr-v100-gupta20a, nagabandi2018neural, xu2020accelerating, kroemer2019review} can learn to cope with hybrid dynamics implicitly, but require large amounts of data to do so, scale poorly as the problem complexity grows, face representational issues near discontinuities, and do not transfer well to significantly different problems. Conversely, hierarchical POMDP planning-based methods \cite{kroemer2019review, jain2018efficient, Brunskill2008, toussaint2008hierarchical} can represent and reason about hybrid dynamics directly, scale well via plan decomposition, work well on novel problems, and reason about uncertainty, but typically rely on precise hand-specified models and task decompositions. We propose a new method, \textbf{M}odel \textbf{I}nference \textbf{C}onditioned on \textbf{A}ctions for \textbf{H}ierarchical Planning (MICAH), that bridges this gap and enables hierarchical POMDP planning-based methods to perform novel manipulation tasks given noisy observations. MICAH infers hybrid automata for objects with configuration-dependent dynamics from unsegmented sequences of observed poses of object parts. These automata can then be used to perform motion planning under uncertainty for novel manipulation tasks involving these objects.

% We combine MICAH with a recently proposed hierarchical POMDP planner, POMDP-HD \cite{jain2018efficient}, and develop a complete pipeline which uses the learned hybrid models to perform motion planning under uncertainty for novel tasks
% Our primary contribution are: (1) an action-conditional inference algorithm for model estimation and changepoint detection from unsegmented data (\textbf{Act}ion conditional \textbf{Ch}angepoint detection using \textbf{A}pproximate \textbf{M}odel \textbf{P}arameters, Act-CHAMP), and (2) an algorithm to construct hybrid automaton for the object using the detected changepoints and the estimated local models. 

MICAH consists of two parts, corresponding to our two main contributions: (1) an novel action-conditional inference algorithm called Act-CHAMP for kinematic model estimation and changepoint detection from unsegmented data, and (2) an algorithm to construct hybrid automata for objects using the detected changepoints and  estimated local models from Act-CHAMP. Due to action-conditional inference, MICAH is more robust to noise and less vulnerable to several modes of failure than existing model inference approaches \cite{niekum2015online, pillai2015learning, martin2019coupled}. These prior approaches assume that the visual pose observations alone provide sufficient information for model estimation, which does not hold for many scenarios and can lead to poor performance. For example, an observation-only approach cannot distinguish between observations obtained by applying force against a rigid object and taking no action at all on a free body, estimating that the model is rigid in both the cases.

% \aj{explain why we need hybrid automata}\aj{benefit: Such a construction ensure that if this model is used with a planner, the planner can plan in continuous space, without having problem of sudden changes in continuous variables.}

To evaluate the proposed method, we first show that for articulated objects, MICAH can correctly infer changepoints and the associated local models with higher fidelity and less data than a state-of-the-art observation-only algorithm, CHAMP \cite{niekum2015online}. We also consider four classes of noisy data to demonstrate its robustness to noise. Next, to test the planning-compatibility of the learned models, we learn hybrid automata for a microwave and a drawer from human demonstrations and use them with a recently proposed hierarchical POMDP Planner, POMDP-HD \cite{jain2018efficient}, to successfully manipulate them in new situations. Finally, we show that the learned models through MICAH are rich-enough to be leveraged creatively by a hierarchical planner for completing novel tasks efficiently---we learn a hybrid automaton for a stapler and use it to dexterously place the stapler at a target point that is reachable only through a narrow corridor in the configuration space.

%%%%%%%%%%%%%%%%%%%%%%%%%%%%%%%%%%%%%%%%%%%%%%%%%%%%%%%%%%%%%%%%%%%%%%%%%%%%%%%%
\section{Related Works}
Learning kinematic models for articulated objects directly from visual data has been studied via different approaches in the literature \cite{sturm2011probabilistic, pillai2015learning, niekum2015online, barragan2014interactive, perez2017c, subramani2018recognizing, hausman2015active, katz2008manipulating,  katz2013interactive, martin2014online, martin2019coupled, abbatematteo2019learning, li2020category}. Sturm et al. \cite{sturm2011probabilistic} proposed a probabilistic framework to learn motion models of articulation bodies from human demonstrations. However, the framework assumes that the objects are governed by a single articulation model, which may not hold true for all objects. For example, a stapler intrinsically changes its articulation state (e.g. rigid vs. rotational) based on the relative angle between its arms. To address this, Niekum et al. \cite{niekum2015online} proposed an online changepoint detection algorithm, \textit{CHAMP}, to detect both the governing articulation model and the temporal changepoints in the articulation relationships of objects. However, all these approaches are observation-only and may fail to correctly infer the object motion model under noisy demonstrations or in cases, when actions are critical for inference.

% In some cases, the provided demonstrations might not be informative enough to estimate the articulation motion model with high confidence. Hausman et al. \cite{hausman2015active} developed an active-learning algorithm to disambiguate the active articulation model of objects by taking informative actions. In another related work, Barrag\'an et al. \cite{barragan2014interactive} proposed a decision-theoretic framework that uses Bayesian filtering to take actions for reducing entropy over the governing model type and its parameters for a mechanism. However, both methods assume that the governing model and its parameters do not change with time. In this work, we remove this assumption and perform model inference even for objects whose motion model changes with time. 

In other closely related works, \textit{interactive perception} approaches aim at leveraging the robot's actions to better perceive objects and build accurate kinematic models \cite{katz2008manipulating, katz2013interactive, martin2014online, martin2019coupled}. Katz et al. first used this approach to learn articulated motion models for planar objects \cite{katz2008manipulating}, and later extended it to use RGB-D data to learn 3D kinematics of articulated objects \cite{katz2013interactive}. Though these approaches use a robot's actions to generate perceptual signals for model estimation, they require the robot's interaction behavior to be pre-scripted by an expert, unlike MICAH, that can estimate models even from noisy demonstrations given by non-expert humans.

An alternative method for learning object motion models is to learn them directly from raw visual data \cite{abbatematteo2019learning, li2020category}. 
% The Embed to Control (E2C) method proposed by Watter et al. \cite{watter2015embed} uses a novel deep probabilistic generative model to convert raw image pixels into a low-dimensional latent space, in which stochastic optimal control can be applied.
% Byravan et al. developed SE3-nets \cite{byravan2017se3} and SE-3Pose-Nets \cite{byravan2018se3} to learn predictive dynamics models of object motion in a scene from input point-cloud data and applied action vectors which can be used to directly perform robot visuomotor control from input point cloud data. 
While deep neural network-based approaches have shown much potential, the biggest hurdle in using such approaches on a wide variety of real-world robotics tasks is the need for a vast amount of training data, which is often not readily available. Also, these approaches tend to transfer poorly to new tasks. In this work, we combine model learning with generalizable planning under uncertainty to address these challenges, though deep learning methods may be useful in future work, in place of our more traditional perception pipeline.

\section{Background}
\subsection{Kinematic Graphs}
\begin{figure}
 \centering
 \includegraphics[width=0.7\linewidth]{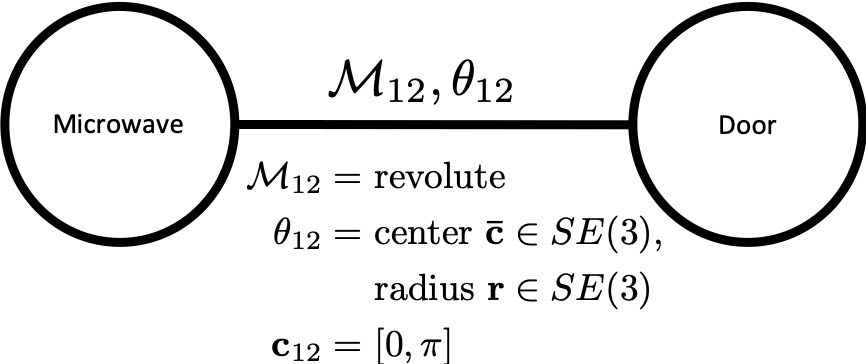}
 \caption{Kinematic graph for microwave which considers the kinematic model as only revolute.}
    \label{fig:kg}
    %  \vspace{-8pt}
\end{figure}

We represent the kinematic structure for articulated objects using \textit{kinematic graphs} \cite{sturm2011probabilistic}. A kinematic graph $G = (V_G, E_G)$ consists of a set of vertices $V_G = {1,...,p}$, corresponding to the $p$ parts of the articulated object, and a set of undirected edges $E_G \subset V_G \times V_G$, each describing the kinematic link between two object parts. An example kinematic graph for a microwave is shown in Figure~\ref{fig:kg}. Sturm et al. \cite{sturm2011probabilistic} proposed to associate a single kinematic link model $\mathcal{M}_{ij}$ with model parameter vector $\theta_{ij}$ with each edge. However, there are many articulated objects with links that are not governed by a single kinematic link model. For example, in most configurations, a microwave door is a revolute joint with respect to the microwave; however, due to the presence of a latch, this relationship changes to a rigid one when the door is closed. In this work, we extend kinematic graphs so that they can represent the hybrid kinematic structure of such objects (see Figure~\ref{fig:ekg}).

\subsection{Changepoint Detection}
\label{sec:champ}
Given a time series of observations $\mathbf{y}_{1:n}$, a changepoint model introduces a number of temporal \textit{changepoints} $\tau_1,...,\tau_m$ that split the data into a set of disjoint segments, with each segment assumed to be governed by a single model (though different models can govern different segments). We build on the online MAP (maximum \textit{a posteriori}) changepoint detection model proposed by Fearnhead and Liu \cite{fearnhead2007line}, which was specialized for detecting motion models for articulated objects by Niekum et al. \cite{niekum2015online}. Given a time series of observations $\mathbf{y}_{1:n}$ and a set of parametric candidate models $M$, the changepoint model infers the MAP set of changepoint times $\boldsymbol{\tau} = \{\tau_0, \tau_1,...,\tau_m, \tau_{m+1} \}$ where $\tau_0 = 0$ and $\tau_{m+1} = n$, giving us $m+1$ segments. Thus, the $k^{th}$ segment consists of observations $\mathbf{y}_{\tau_{k}+1 : \tau_{k+1}}$, and has an associated model $\mathcal{M}^k \in \mathbb{M}$ with parameters $\theta^k$.

Assuming that the data after a changepoint is independent of the data prior to that changepoint, we model the position of changepoints in the time series as a Markov chain in which the transition probabilities are defined by the time since the last changepoint,
\begin{equation}
    p(\tau_{i+1} = t | \tau_i = j) = \beta(t-j)
    \label{eq:eqnPre3}
\end{equation}
where $\beta(\cdot)$ is a probability distribution over time. For a segment from time $s$ to $t$, the model evidence for the governing model being $\mathcal{M}$, is defined as:
\begin{equation}
 L(s,t,\mathcal{M}) = p(\mathbf{y}_{s+1:t} | \mathcal{M}) = \int p(\mathbf{y}_{s+1:t} | \mathcal{M}, \theta) p(\theta) d\theta
\end{equation}{}
The distribution over the position of the most recent changepoint prior to time t, $C_t$, can be efficiently estimated using the standard Bayesian filtering recursions and an online Viterbi algorithm \cite{fearnhead2007line}. We define $E_s$ as the event that given a changepoint at time $s$, the MAP choice of changepoints has occurred prior to time $s$. Then, the probability of having a changepoint at time $t$, $P_t$, is defined as:
\begin{align}
    \begin{split}
        P_t &= p(C_{t} = s, \mathcal{M}, E_s, \mathbf{y_{1:t}}) \\
        P^{MAP}_t &= p(\text{Changepoint at t}, E_s, \mathbf{y_{1:t}})
    \end{split}
\end{align}
which results in 
\begin{align}
    \begin{split}
        P_t(s,\mathcal{M}) &= (1- B(t-s-1))~L(s,t, \mathcal{M})~p(\mathcal{M})~P^{MAP}_s \\
        P^{MAP}_t &= \max_{s, \mathcal{M}} \left[ \frac{\beta(t-s)}{1- B(t-s-1)}P_t(s, \mathcal{M}) \right]
    \end{split}
    \label{eq:eqnPre2}
\end{align}
where $B(\cdot)$ is the cumulative distribution function of $\beta(\cdot)$. By finding the values of $(s, \mathcal{M})$ that maximize $P_t^{MAP}$, the Viterbi path can be recovered at any point. This process can be repeated until the time $t=0$ is reached to estimate all changepoints that occurred in the given time series $\mb{y}_{1:T}$.

The algorithm is fully online, but requires $O(n)$ computations at each time step, since $P_t(s,\mathcal{M})$ values must be calculated for all $s < t$. The computation time is reduced to a constant by using a particle filter that keeps a constant number of particles, $M$, at each time step, each of which represents a support point in the approximate density $p(C_t=s, \mathbf{y}_{1:t})$. If at any time step, the number of particles exceeds $M$, stratified optimal resampling \cite{fearnhead2007line} is used to choose which particles to keep such that the Kolmogorov-Smirnov distance from the true distribution is minimized in expectation.

\subsection{Hybrid Automaton}
A hybrid automaton describes a dynamical system which evolves both in the continuous space and over a finite set of discrete states with time \cite{lygeros2012hybrid}. Formally, a hybrid automaton is a collection $\mathsf{H} = (Q, X, U, Init, f, I, \mathcal{E}, \mathcal{G}, R, \phi)$, where
% \begin{itemize}
%     \item $Q$ is a finite set of discrete states;
%     \item $X = \mathbb{R}^N$ is a set of continuous states;
%     \item $U = U_D \cup U_C$ is a finite collection of input variables, where $U_D$ contain discrete and $U_C$ contain continuous variables;
%     \item $Init \subseteq \textbf{Q} \times \textbf{X}$ is a set of initial states;
%     \item $f:\textbf{Q} \times \textbf{X} \times \textbf{U} \rightarrow \mathbb{R}^N$ is an input dependent vector field that defines the evolution of the continuous states with time;
%     \item $I:\textbf{Q} \rightarrow 2^{\textbf{X} \times \textbf{U}}$ assigns to each $q \in \textbf{Q}$ an input dependent invariant set;
%     \item $\mathcal{E} \subseteq \textbf{Q} \times \textbf{Q}$ is a collection of discrete transitions (edges);
%     \item $\mathcal{G}: \mathcal{E} \rightarrow 2^{\textbf{X}\times \textbf{U}}$ assigns to each $e=(q, q') \in \mathcal{E}$ a guard;
%     \item $R: \mathcal{E} \times \textbf{X} \times \textbf{U} \rightarrow 2^{X}$ assigns to each $e=(q, q') \in \mathcal{E}$, $x \in \textbf{X}$, and $u \in \textbf{U}$ a reset relation; and,
%     \item $\phi : \textbf{Q}\times \textbf{X} \rightarrow 2^{\textbf{U}}$ assigns to each state $q \in \textbf{Q}$ a set of admissible inputs.
% \end{itemize}{}
each discrete state $q \in Q$ of the system can be interpreted as representing a separate local dynamics model $f:\textbf{Q} \times \textbf{X} \times \textbf{U} \rightarrow \mathbb{R}^N$ that governs the evolution of the continuous states $x \in X = \mathbb{R}^N$ under applied actions $u \in U$. $Init \subseteq \textbf{Q} \times \textbf{X}$ represents the set of initial states. The discrete state transitions can be represented as a directed graph with each possible discrete state $q$ corresponding to a node and edges ($e \in \mathcal{E} \subseteq \textbf{Q} \times \textbf{Q}$) marking possible transitions between the nodes. These transitions are conditioned on the continuous states through guards $\mathcal{G}: \mathcal{E} \rightarrow 2^{\textbf{X}\times \textbf{U}}$. A transition from the discrete state $q$ to another state $q'$ happens if the continuous states $\mathbf{x}$ are in the $\mathcal{G}(e)$ of the edge $e = (q, q') \in \mathcal{E}$. 
$I:\textbf{Q} \rightarrow 2^{\textbf{X} \times \textbf{U}}$ assigns to each $q \in \textbf{Q}$ an input dependent invariant set $I(q)$, such that $\forall x_0 \in I(q) \subseteq X$ there exists a control law $u_t = h(x_t)$ such that $x_t \in I(q)$ for all $t \geq 0$ and $u_t = h(x_t) \in U$. 
Reset map $R: \mathcal{E} \times \textbf{X} \times \textbf{U} \rightarrow 2^{X}$ assigns to each $e=(q, q') \in \mathcal{E}$, $x \in \textbf{X}$, and $u \in \textbf{U}$ a map that relates the values of the continuous states before and after the discrete state transition through the edge $e$. The set of admissible inputs for each state $q \in \textbf{Q}$ is defined using $\phi: \textbf{Q}\times \textbf{X} \rightarrow 2^{\textbf{U}}$. 

% A finite state Markov chain can be used to model the evolution of the discrete states of the system. Defining the state transition matrix as $\Pi = \{\pi_{ij} \}$, the discrete state evolution can be given as $q_{t+1} = \Pi q_t$.

% Thus, for each discrete state $q$, in a hybrid dynamics model we can define $~\mb{x}_{t+1} = F^{q}(\mb{x}_t,\mb{u}_t),~ \mb{z}_t = H^{q}(\mb{x}_t)$ where $\mb{x} \in \mathbb{R}^n$, $\mb{u} \in \mathbb{R}^m$, $\mb{z} \in \mathbb{R}^l$, $F^q(\mb{x}, \mb{u})$ and $H^q(\mb{x})$ are the continuous state, control input, observation variables, state dynamics and observation functions respectively.

%%%%%%%%%%%%%%%%%%%%%%%%%%%%%%%%%%%%%%%%%%%%%%%%%%%%%%%%%%%%%%%%%%%%%%%%%%%%%%%%
% \begin{figure*}
%   \begin{subfigure}[b]{0.48\textwidth}
%     \includegraphics[width=0.8\textwidth]{figures/kinematic_graphs.png}
%     \caption{Kinematic graph for microwave which considers the kinematic model as only revolute.}
%     \label{fig:kg}
%   \end{subfigure}
%   \hfill 
%   %
%   \begin{subfigure}[b]{0.48\textwidth}
%     \includegraphics[width=0.9\textwidth]{figures/extended_kinematic_graphs.png}
%     \caption{Extended kinematic graph for microwave which considers a hybrid model that can be revolute or rigid, depending on the configuration.}
%     \label{fig:ekg}
%   \end{subfigure}
% \end{figure*}

\begin{figure}
 \centering
 \includegraphics[width=0.85\linewidth]{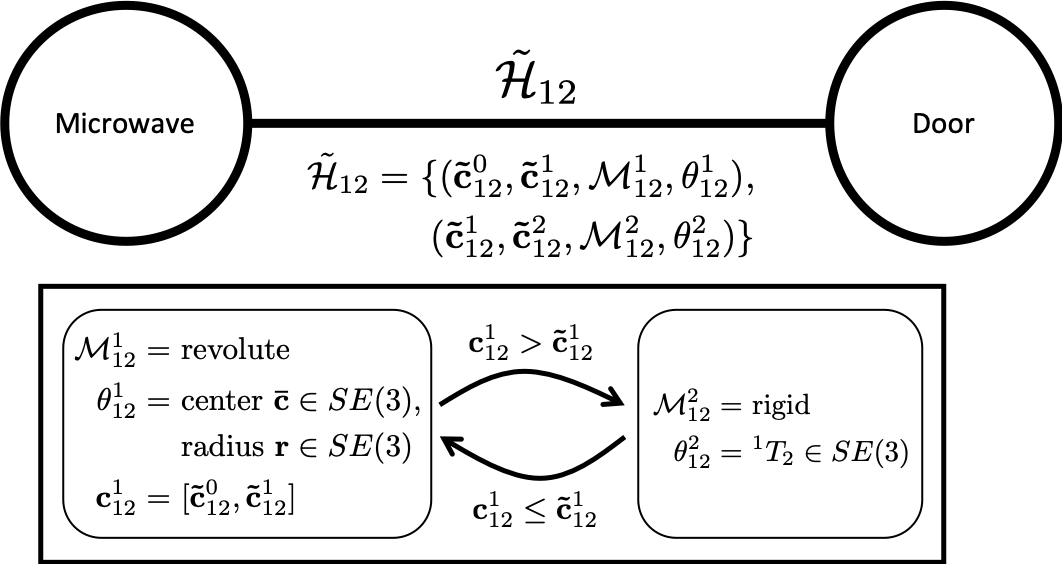}
   \caption{Extended kinematic graph for microwave which considers a hybrid model that can be revolute or rigid, depending on the configuration.}
    \label{fig:ekg}
    % \vspace{-8pt}
\end{figure}

\section{Approach}
Given a sequence of object part pose observations $\mathcal{D}_{\mathbf{y}}$ (e.g. from a visual tracking algorithm) and a sequence of applied actions $\mathcal{D}_{\mathbf{a}}$ on an articulated object, MICAH creates a planning-compatible hybrid automaton for the object. It does so in two steps: (1) it estimates the kinematic graph $\hat{G}$ representing the kinematic structure of the object given the sequence of pose observations $\mathcal{D}_{\mathbf{y}}$ and the applied actions $\mathcal{D}_{\mathbf{a}}$, and then (2) constructs a hybrid automaton $\mathsf{H}$ representing the motion model for the object given $\hat{G}$. 

For the first step, we extend the framework proposed by Sturm et al. \cite{sturm2011probabilistic} in two important ways to better learn the kinematic structure of articulated objects. First, we include reasoning about the applied actions along with the observed motion of the object while estimating its kinematic structure. Second, we extend the framework to be able to learn the kinematic structure of more complex articulated objects that may exhibit configuration-dependent kinematics, e.g., a microwave. The original framework \cite{sturm2011probabilistic} assumes that each link of an articulated body is governed by a single kinematic model. For complex articulated objects that exhibit configuration-dependent kinematics, the transitions points in the kinematic model along with the set of governing local models and their parameters need to be estimated to learn the complete kinematic structure of the object. 

To facilitate these extensions, we introduce a novel action-conditional changepoint detection algorithm, \textbf{Act}ion conditional \textbf{Ch}angepoint detection using \textbf{A}pproximate \textbf{M}odel \textbf{P}arameters (Act-CHAMP), that can detect the changepoints in the relative motion between two rigid objects (or two object parts), given a time series of observations of the relative motion between the objects and the corresponding applied actions. The algorithm is described in section~\ref{sec:actCHAMP}.

Kinematic trees have the property that their edges are independent of each other. As a result, when learning the kinematic relationship between object parts $i$ and $j$ of an articulated object, only their relative transformations are relevant for estimating the edge model. MICAH first uses the Act-CHAMP algorithm to learn the kinematic relationships between different parts of the articulated object separately, and then combines them to estimate the complete kinematic graph $\hat{G}$ for the object. Once the kinematic graph $\hat{G}$ for an articulated object is known, MICAH constructs a hybrid automaton $\mathsf{H}$ to represent its motion model. We choose hybrid automata as they present a natural choice to model the motion of objects that may exhibit different motion models based on their configuration. Steps to construct a hybrid automation from the learned kinematic graph $\hat{G}$ is described in section~\ref{sec:hybridModeling}.

\subsection{Action-conditional Model Inference}
\label{sec:actCHAMP}
Following Sturm et al. \cite{sturm2011probabilistic}, we define the relative transform between two objects with poses $x_i$ and $x_j \in SE(3)$ at time $t$ as: $~\boldsymbol{\Delta}_{ij, t} = \mathbf{x}_{i,t} \ominus \mathbf{x}_{j,t}$\footnotemark[1].\footnotetext[1]{The operators $\oplus$ and $\ominus$ represent motion composition operations. For example, if poses $\mathbf{x}_i$, $\mathbf{x}_j \in \mathcal{R}^{4\times4}$  are represented as homogeneous matrices, then these operators correspond to matrix multiplications $\mathbf{x}_i \oplus \mathbf{x}_j  = \mathbf{x}_i \mathbf{x}_j$ and its inverse multiplication, $\mathbf{x}_i \ominus \mathbf{x}_j  = \mathbf{x}_i^{-1} \mathbf{x}_j$, respectively.} Additionally, we define an action $\mathbf{a}_t$ taken by the demonstrator at a time $t$ as the intended displacement to be applied to the relative transform between two objects from time $t$ to $t+1$ as:~~$\mathbf{a}_t = \boldsymbol{\Delta}_{ij, t} \ominus \boldsymbol{\Delta}_{ij, t+1}$. Given the time-series of observations $\mathcal{D}_{\mathbf{y}_{ij}} = \mathbf{y}_{1: T}$ of relative motion between the two object parts $i$ and $j$ of an articulated object and the corresponding applied actions $\mathcal{D}_{\mathbf{a}_{ij}} = \mb{a}_{1:T}$, we wish to find the set $\mathcal{\tilde{H}}_{ij}$ defining the kinematic relationship between the two object parts. The set $\mathcal{\tilde{H}}_{ij}$ consists of tuples $(\mathbf{\tilde{c}}^{k-1}_{ij}, \mathbf{\tilde{c}}^{k}_{ij}, \mathcal{M}^{k}_{ij}, \theta^{k}_{ij})$, where $\mathbf{\tilde{c}}^{k-1}_{ij}$ and $\mathbf{\tilde{c}}^{k}_{ij}$ denote the starting and the end configurations for the model $\mathcal{M}^k_{ij}$ to be the governing local model with parameters $\theta^k_{ij}$. For the sake of clarity, we drop the subscript $\{ij\}$ in the following discussion in this section.

We propose a novel algorithm, Act-CHAMP, that performs action-conditional changepoint detection to estimate the set $\tilde{\mathcal{H}}_{ij}$ given input time series of observations $\mathbf{y}_{1:T}$ and the corresponding applied actions $\mathbf{a}_{1:T}$. Act-CHAMP builds upon the CHAMP algorithm proposed by Niekum et al. \cite{niekum2015online}. The CHAMP algorithm reasons only about the observed relative motion between the objects for estimating the kinematic relationship between the objects. However, an observation-only approach can easily lead to false detection of changepoints and result in an inaccurate system model. Consider an example case of deducing the motion model for a drawer from a noisy demonstration in which the majority of applied actions are orthogonal to the axis of motion of the drawer. Due to intermittent displacements, an observation-only approach might model the motion of the drawer to be comprised of a sequence of multiple rigid joints. On the other hand, an action-conditional inference can maintain an equal likelihood of observing either a rigid or a prismatic model under off-axis actions, leading to a more accurate model. 

Given the two time series inputs $\mathbf{y}_{1:T}$ and $\mathbf{a}_{1:T}$, we define the model evidence for model $\mathcal{M}$ being the governing model for the time segment between times $s$ and $t$ as:
\begin{align}
    \begin{split}
        L(s, t, \mathcal{M}) &= p(\mb{y}_{s+1:t}|\mathcal{M}, \mb{a}_{s:t-1}) \\
    &= \int p(\mb{y}_{s+1:t}|\mathcal{M}, \theta, \mb{a}_{s: t-1}) p(\theta) d\theta
    \label{eq:eqn5}
    \end{split}
\end{align}{}

Each model $\mathcal{M}$ admits two functions: a forward kinematics function, $f_{\mathcal{M}, \theta}$, and an inverse kinematics function, $f^{-1}_{\mathcal{M}, \theta}$, which maps the relative pose between the objects $\boldsymbol{\Delta}_{ij}$ to a unique configuration $\mathbf{c}$ for the model (e.g. a position along the prismatic axis, or an angle with respect to the axis of rotation) as:
\begin{align*}
\begin{split}
  f_{\mathcal{M}, \theta}(\mathbf{c}_{\mathcal{M}}) = \boldsymbol{\Delta} ~~~~~&\text{(forward kinematics)} \\
   f^{-1}_{\mathcal{M}, \theta}(\boldsymbol{\Delta}) = \mathbf{c}_{\mathcal{M}} ~~~~~&\text{(inverse kinematics)}
\end{split}{}
\end{align*}
We consider three candidate models $\mathcal{M}^{\mathrm{rigid}}$, $\mathcal{M}^{\mathrm{revolute}}$, and $\mathcal{M}^{\mathrm{prismatic}}$ to define the kinematic relationship between two objects. Complete definitions of forward and inverse kinematics models for these models are beyond the scope of this work; for more details, see Sturm et al. \cite{sturm2011probabilistic}. 

Additionally, we define the Jacobian and inverse Jacobian functions for the model $\mathcal{M}$ as 
\begin{align*}
\begin{split}{}
 J_{\mathcal{M}, \theta}(\delta{\mathbf{c}_{\mathcal{M}}}) = \delta{\boldsymbol{\Delta}} ~~~~~~&\text{(Jacobian)} \\
  J^{-1}_{\mathcal{M}, \theta}(\delta{\boldsymbol{\Delta}}) = \delta{\mathbf{c}_{\mathcal{M}}} ~~~~~&\text{(inverse Jacobian)}
\end{split}
\end{align*}{}
where $\delta \boldsymbol{\Delta}$ and $\delta \mathbf{c}_{\mathcal{M}}$ represent small perturbations applied to the relative pose and the configuration, respectively. 

Using these functions, we can define the likelihood of obtaining observations $\mathbf{y}_{1:T}$ upon applying action $\textbf{a}_{1:T}$  under model $\mathcal{M}$ as:
\begin{equation}
    p(\mathbf{y}_{2:T}|\mathcal{M}, \theta, \textbf{a}_{1:T-1}) = \prod^T_{t=2} p(\mb{y}_t \mid \hat{\mathbf{\Delta}}_{t})
    \label{eq:eqn1}
\end{equation}
where $\hat{\mathbf{\Delta}}_{t}$ is the predicted relative pose under the model $\mathcal{M}$ at time t, and can be calculated using the observation $\mathbf{y}_{t-1}$ and applied action $\mathbf{a}_{t-1}$ at time $t-1$ as:
\begin{equation}
    \hat{\boldsymbol{\Delta}}_{t} = f_{\mathcal{M}, \theta}(~f^{-1}_{\mathcal{M}, \theta}(\mathbf{y}_{t-1}) + J_{\mathcal{M}, \theta}^{-1}~\mathbf{a}_{t-1})
\label{eq:eqn2}
\end{equation}

The probability $p(\mb{y}_t \mid \hat{\mathbf{\Delta}}_{t})$ can be calculated by defining an observation model, given an observation error covariance $\Sigma_y$ for the perception system as:
\begin{align}
\mathbf{y}_t \sim 
\begin{cases}
\boldsymbol{\Delta}_t + \mathcal{N}(0, \Sigma_y)~~~~&\text{if $\nu$ = 1} \\
\mathcal{U}~~~&\text{if $\nu$ = 0}
\end{cases}
\end{align}{}
where the probability of observation being an outlier is $p(\nu = 0) = \gamma$, in which case it is drawn from a uniform distribution $\mathcal{U}$. The data likelihood is then defined as:
\begin{gather}
    p(\mathbf{y}_t| \boldsymbol{\Delta}_t) = p(\mathbf{y}_t|\boldsymbol{\Delta}_t, \gamma)p(\gamma), ~~~~\text{where,} \\
        p(\mathbf{y}_t|\boldsymbol{\Delta}_t, \gamma) = (1-\gamma)p(\mathbf{y}_t|\nu=1) + \gamma p(\mathbf{y}_t|\nu=0),\\
        p(\gamma) \propto e^{-w\gamma},
\end{gather}
and $w$ is a weighting constant.

Finally, similar to Niekum et al. \cite{niekum2015online}, we can define our BIC-penalized likelihood function as:
\begin{equation}
    \ln L(s, t, \mathcal{M}) \approx \ln p(\mb{y}_{s+1:t}|\mathcal{M}, \hat{\theta}, \mb{a}_{s:t-1}) - \frac{1}{2}k_q \ln (t-s)
    \label{eq:me1}
\end{equation}{}
where estimated parameters $\hat{\theta}$ are inferred using MLESAC (Maximum Likelihood Estimation Sample Consensus) \cite{torr2000mlesac}. This likelihood function can be used in conjugation with the changepoint detection algorithm described in section~\ref{sec:champ} to infer the MAP set of changepoint times $\boldsymbol{\tau}$ along with the associated local models $\mathcal{M}^k \in \mathbb{M}$ with parameters $\theta^k$. The detected changepoints $\boldsymbol{\tau}$ and the local models can be later combined appropriately to obtain a set $\mathcal{H}_{ij}$ consisting of tuples $(\tau^{k-1}_{ij}, \tau^{k}_{ij}, \mathcal{M}^{k}_{ij}, \theta^{k}_{ij})$, where $\tau^{k-1}_{ij}$ and $\tau^{k}_{ij}$ denote the starting and the end changepoints for the time segment $k$ in the input time series $\mathbf{y}_{1:T}$.

The transition conditions between the local models can be made independent of the changepoint times, $\boldsymbol{\tau}$, by making use of the observations corresponding to the changepoint times $\mathbf{y}_{\mathbf{\tau}} \subseteq \mathbf{y}_{1:T}$. If an observation $\mathbf{y}_{\tau^{k}}$ corresponds to the changepoint $\tau^{k}$ denoting the transition from local model $\mathcal{M}^{k}$ to the model $\mathcal{M}^{k+1}$, then the inverse kinematics function $f^{-1}_{\mathcal{M}^{k},\theta^{k}}$ can be used to find an equivalent configurational changepoint $\mathbf{\tilde{c}}^{k}$, a fixed configuration for model $\mathcal{M}^{k}$, that marks the transition from model $\mathcal{M}^{k}$ to the next model $\mathcal{M}^{k+1}$. We can thus convert the set $\mathcal{H}_{ij}$ to the set $\tilde{\mathcal{H}}_{ij}$, consisting of tuples $(\mathbf{\tilde{c}}^{k-1}_{ij}, \mathbf{\tilde{c}}^{k}_{ij}, \mathcal{M}^{k}_{ij}, \theta^{k}_{ij})$, that is independent of the input time series.

The complete kinematic structure of the articulated object can then be estimated by finding the set of edges $E_G$, denoting the kinematic connections between its parts, that maximizes the posterior probability of observing $\mathcal{D}_z$ under applied actions $\mathcal{D}_a$ \cite{sturm2011probabilistic}. However, to account for complex articulated objects that exhibit configuration-dependent kinematics, now each edge $\tilde{e}_{ij} \in E_G$ of the kinematic graph $G$ can correspond to multiple kinematic link models $\mathcal{M}^k_{ij}$, unlike the original framework \cite{sturm2011probabilistic}, in which each edge corresponds to only one kinematic link model $\mathcal{M}_{ij} \in \mathbb{M}_{ij}$. To denote the change, we call such kinematic graphs, \textit{extended kinematic graphs}. An example extended kinematic graph for a microwave is shown in Figure~\ref{fig:ekg}. 

% \begin{align}
%         \hat{E}_G &= \argmax_{E_G} p(E_G | \mathcal{D}_z, \mathcal{D}_a) \label{eq:ekg1}\\
%         &= \argmax_{E_G} p(\{ \hat{\tilde{\mathcal{H}}}_{ij} | \tilde{e}_{ij} \in E_G\} | \mathcal{D}_z, \mathcal{D}_a) \label{eq:ekg2}\\
%         &= \argmax_{E_G} \prod_{\tilde{e}_{ij} \in E_G} p(\hat{\tilde{\mathcal{H}}}_{ij} | \mathcal{D}_{z_{ij}}, \mathcal{D}_{a_{ij}}) \label{eq:ekg3}\\
%         &= \argmax_{E_G} \sum_{\tilde{e}_{ij} \in E_G} \log  p(\hat{\tilde{\mathcal{H}}}_{ij} | \mathcal{D}_{z_{ij}}, \mathcal{D}_{a_{ij}}) \label{eq:ekg4}
% \end{align}
% where, for going from Equation~\ref{eq:ekg2} to Equation~\ref{eq:ekg3}, we leverage the property of kinematic graphs that their individual edges are independent of each other \cite{sturm2011probabilistic}. The kinematic tree that maximizes (\ref{eq:ekg4}) corresponds to the minimum spanning tree in a fully connected graph with edge costs:
% \begin{align}
%     \text{cost}_{ij} &= -\log p(\hat{\tilde{\mathcal{H}}}_{ij} | \mathcal{D}_{z_{ij}}, \mathcal{D}_{a_{ij}}) \\
%     &= - \sum^{m+1}_{k=1} \log p((\mathbf{\tilde{c}}^{k-1}_{ij}, \mathbf{\tilde{c}}^{k}_{ij}, \hat{\mathcal{M}}^k_{ij}, \hat{\theta}^k_{ij})\}| \mathcal{D}_{z_{ij}}, \mathcal{D}_{a_{ij}})
% \end{align}{}
% that we approximate using the BIC value. The best kinematic structure can now be found efficiently, in $\mathcal{O}(p^2\log p)$ time, using Prim's or Kruskal's algorithm for finding the minimum spanning trees \cite{sturm2011probabilistic, cormen2009introduction}.

\subsection{Hybrid Automaton Construction}
\label{sec:hybridModeling}
Hybrid automata present a natural choice for representing an articulated object that can have a discrete number of configuration-dependent kinematics models. A hybrid automaton can model a system that evolves over both discrete and continuous states with time effectively, which facilitates robot manipulation planning for tasks involving that object. We define the hybrid automaton $\mathsf{H} = (Q, X, U, Init, f, I, \mathcal{E}, \mathcal{G}, R, \phi)$ for the articulated object as:

\begin{itemize}
    \item $Q = \prod_{\tilde{e}_{ij} \in E_G} \mathbb{M}_{ij}$, i.e. the Cartesian product of the sets of local models defining kinematic relationship between two object parts;
    \item $X = \prod_{\tilde{e}_{ij} \in E_G} \bar{c}_{ij}$, where we use a single variable $\bar{c}_{ij} \in \mathbb{R}$ to represent the configuration value $\mathbf{c}_{\mathcal{M}^k}$ under all models $\mathcal{M}^k_{ij} \in \mathbb{M}_{ij}$, as each of the candidate articulation models admits a single-dimensional configuration variable $\mathbf{c}_{\mathcal{M}} \in \mathbb{R}$;
    \item $U = U_C = \prod_{\tilde{e}_{ij} \in E_G} u_{ij}$, where $u_{ij} \in \mathbb{R}$ is the input delta to be applied to the continuous state $\bar{c}_{ij}$ and the set of discrete input variables is the null set $U_D = \emptyset$ as we cannot control the discrete states directly;
    \item $Init$ is defined as per the task definition;
    \item The vector field $f$ governing the evolution of the continuous state vector $x$ with time is defined as $f(q, x, u) = (\mathbf{x}_t - \mathbf{x}^q ) + \mathbf{u}_t$, where $q \in Q$, $\mathbf{x}_t , \mathbf{x}^q\in X$, and $\mathbf{u} \in U$. The vector $\mathbf{x}^q \in X$ is so defined that its $l$-th element $\mathbf{x}^q[l] = \sum^{k-1}_{r=0} \mathbf{\tilde{c}}^{r}_{ij}$, where $l$-th dimension of $X$ corresponds to the kinematic relationship between object parts $i$ and $j$ with $\tilde{e}_{ij} \in E_G$, and $q[l] = \mathcal{M}^k_{ij}$;
    \item For each discrete state $q \in \textbf{Q}$, an invariant set $I(q)$ is defined such that within it the time evolution of the continuous states is governed by the vector field $f(q, x, u)$ $\forall x \in I(q) \subseteq X, u \in U$. We define $I$ as $I = \prod_{\tilde{e}_{ij} \in E_G} Dom(\mathbb{M}_{ij})$, where $Dom(\mathbb{M}_{ij}) = \{Dom(\mathcal{M}^k_{ij}) \forall k \in |\mathbb{M}| \}$ with $Dom(\mathcal{M}^k_{ij})$ defined as $Dom(\mathcal{M}^k_{ij}) = [0, \mathbf{\tilde{c}}^{k+1}_{ij})$;
    \item The set of edges defines the set of feasible transitions between the discrete states, $\mathcal{E} = \{(q, q')~|~ q[l]=\mathcal{M}^{k}_{ij} \Rightarrow q'[l]=\mathcal{M}^{r}_{ij}, r\in\{k-1, k, k+1\}\}$;
    \item Guards $\mathcal{G}$ can be constructed using the configurational changepoints estimated for the object. If an edge $e = (q, q') \in \mathcal{E}$ corresponds to a transition from a local model $\mathcal{M}^k_{ij}$ to model $\mathcal{M}^{k+1}_{ij}$, then the guard for the edge $e$ can be defined as $\mathcal{G}(e) = \{\bar{c}_{ij} \geq \tilde{\mathbf{c}}^{k+1}_{ij}\}$. Analogously, the guard for the reverse transition $\mathcal{G}(e'=(q',q)) = \{\bar{c}_{ij} < 0\}$. To handle the corner cases when $\bar{c}_{ij} < 0$ for $\mathcal{M}^1_{ij}$ or $\bar{c}_{ij} > \mathbf{\hat{c}}^{m+1}_{ij}$ for model $\mathcal{M}^{m}_{ij}$ (assuming $|\mathbb{M}_{ij}| = m$), we define two additional edges $e^0_{ij}$ and $e^{-1}_{ij}$ which corresponds to the self transitions to the same discrete states such that $\bar{c}_{ij}$ is lower-bounded at $0$ for $\mathcal{M}^1_{ij}$ and upper-bounded at $ \mathbf{\hat{c}}^{m+1}_{ij}$ for model $\mathcal{M}^m_{ij}$; 
    \item The reset map $R$ is an identity map;
    \item The set of admissible inputs $\phi(q, \mathbf{x}) = \mathbf{U}$.
\end{itemize}

\section{Experiments and Discussions}
In the first set of experiments, we compare the performance of Act-CHAMP with the CHAMP algorithm \cite{niekum2015online} to estimate changepoints and local motion models for a microwave and a drawer. Next, we test the complete method, MICAH, to construct planning-compatible hybrid automata for the microwave and drawer and discuss the results of manipulation experiments to open and close the microwave door and the drawer using the learned models. Finally, we show that MICAH can be combined with a recent hierarchical POMDP planner, POMDP-HD \cite{jain2018efficient}, to develop a complete pipeline that can learn a hybrid automaton from demonstrations and leverage it to perform a novel manipulation task---in this case, with a stapler. A video showcasing the experiments is available at: {\small \color{blue} \url{https://youtu.be/f35gMoOoOy8}}.

\subsection{Learning Kinematics Models for Objects}
We collected six sets of demonstrations to estimate motion models for the microwave and the drawer. We provided kinesthetic demonstrations to a two-armed robot, in which the human expert physically moved the right arm of the robot, while the left arm shadowed the motion of the right arm to interact with objects while collecting unobstructed visual data. The first two sets provide low-noise data, by manipulating the door handle or drawer knob via a solid grasp. The next two sets provide data in which random periods of no actions on the objects were deliberately included while giving demonstrations. The last two sets consist of high-noise cases, in which the actions were applied by pushing with the end-effector without a grasp. Relative poses of object parts were recorded as time-series observations with an RGB-D sensor using the SimTrack object tracker \cite{pauwels_simtrack_2015}. For each time step $t$, the demonstrator's action $\mathbf{a}_t$ on the object was defined as the difference between the position of the right end-effector at times $t$ and $t+1$.  

\textbf{With grasp:} % \label{microwave} 
Both algorithms (CHAMP and Act-CHAMP) detected a single changepoint in the articulated motion of the microwave door and determined the trajectory to be composed of two motion models, namely rigid and revolute. For the drawer, both algorithms were able to successfully determine its motion to be composed of a single prismatic motion model(see Table~\ref{table:table1}). This demonstrates that for clean, information-rich demonstrations, Act-CHAMP can perform on par with the baseline.

%\begin{figure}[b]
%    \centering
%    \subfloat[Microwave]{\includegraphics[height=0.30\linewidth, width=0.3\linewidth]{figures/microwave_action.png}} ~~~~~~~~~~%
%    \subfloat[Drawer]{\includegraphics[height=0.3\linewidth, width=0.3\linewidth]{figures/drawer_w_grip_action.png}}
%    \caption{Inferred motion models for the microwave and the drawer using Act-CHAMP. Points denote the recorded relative poses of object parts from one demonstration. The small solid circle represents the detected rigid model, the circular arc represents the detected revolute model, and the line represents the detected prismatic model.}%
%    \label{fig:figure1}%
%    %  \vspace{-5pt}
%\end{figure}

\begin{figure}[b]
	\centering
	\begin{subfigure}{0.3\linewidth}
	\centering\includegraphics[height=\linewidth, width=\linewidth]{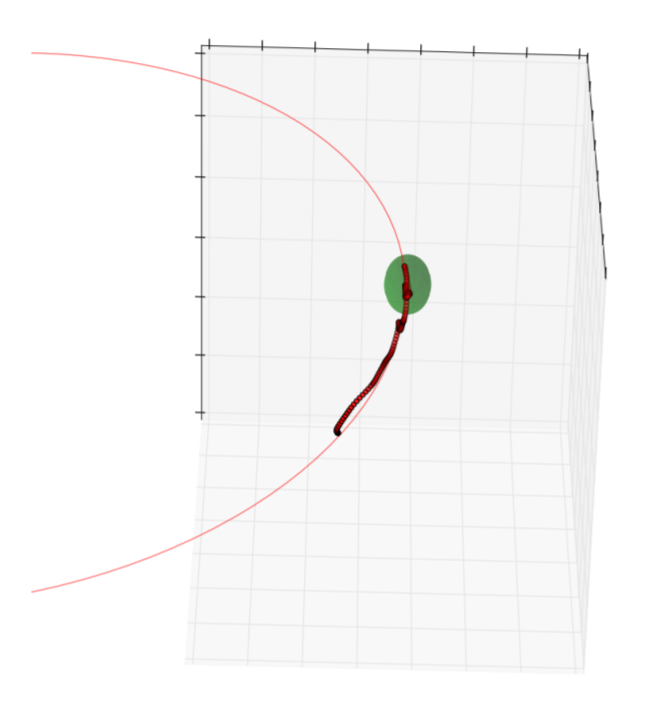}
	\caption{Microwave}
	\end{subfigure} \qquad
	\begin{subfigure}{0.3\linewidth}
		\centering\includegraphics[height=0.95\linewidth, width=0.95\linewidth]{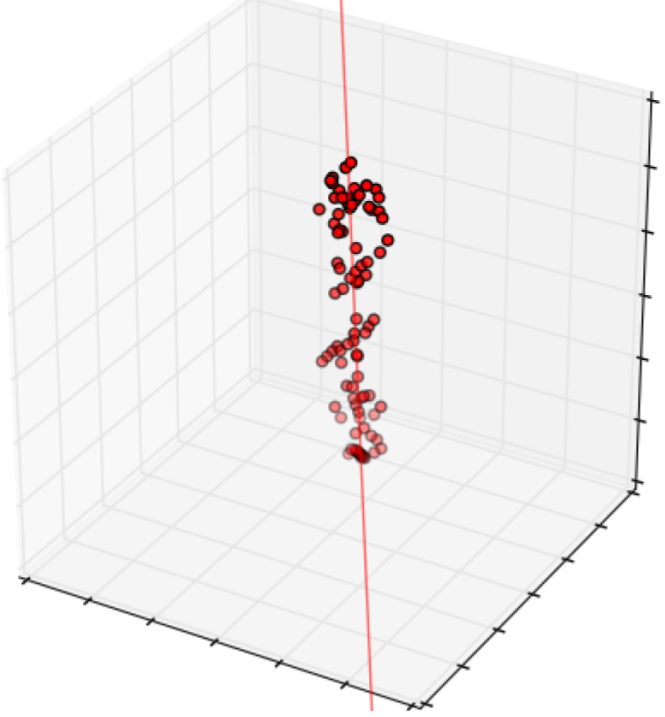}
		\caption{Drawer}
	\end{subfigure}
	\caption{Inferred motion models for the microwave and the drawer using Act-CHAMP. Points denote the recorded relative poses of object parts from one demonstration. The green circle represents the detected rigid model, the circular arc represents the detected revolute model, and the line represents the detected prismatic model.}
	\label{fig:figure1}
\end{figure}

%\begin{figure}[b]
%    \centering
%    \subfloat[CHAMP]{ \includegraphics[height=0.33\linewidth, width=0.35\linewidth]{figures/drawer_wo_grip_champ.png}} ~~~~~~~~%
%    \subfloat[Act-CHAMP]{\includegraphics[height=0.3\linewidth, width=0.35\linewidth]{figures/drawer_w_grip_action.png}}%
%    \caption{Act-CHAMP correctly infers the drawer motion model, while CHAMP (baseline) falsely detects a changepoint under noisy demonstrations.}%    
%    \label{fig:figure2}%
%    % \vspace{-5pt}
%\end{figure}

\begin{figure}[b]
	\centering
	\begin{subfigure}{0.35\linewidth}
		\includegraphics[height=0.9\linewidth, width=0.9\linewidth]{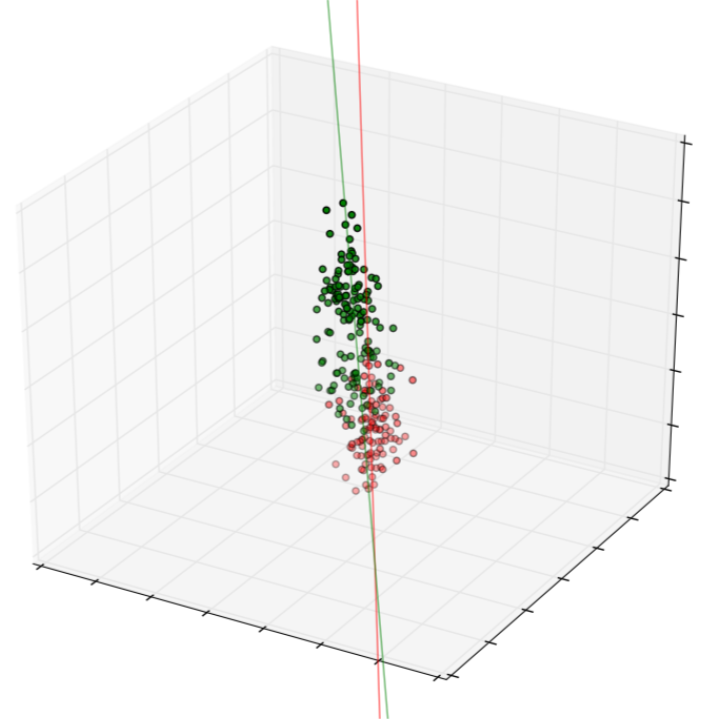}
		\caption{CHAMP}
	\end{subfigure} \qquad
	\begin{subfigure}{0.33\linewidth}
		\includegraphics[height=0.9\linewidth, width=0.9\linewidth]{figures/drawer_w_grip_action.png}
		\caption{Act-CHAMP}
	\end{subfigure}
	\caption{Act-CHAMP correctly infers the drawer motion model, while CHAMP (baseline) falsely detects a changepoint under noisy demonstrations.}%    
	    \label{fig:figure2}%
\end{figure}

\textbf{No-Actions:}
When no action is applied to an object, due to the lack of motion, an observation-only model inference algorithm can infer the object motion model to be rigid. Moreover, if the agent stops applying actions after interacting with the object for some time, an observation-only approach can falsely detect a changepoint in the motion model. We hypothesize that an action-conditional inference algorithm such as Act-CHAMP won't suffer from these shortcomings as it can reason that no motion is expected if no actions are applied. To test it, we conducted experiments in which the demonstrator stopped applying actions on the object midway during a demonstration for an extended time randomly at two distinct locations. As expected, the observation-only CHAMP algorithm falsely detected changepoints in the object motion model and performed poorly (see Table~\ref{table:table1}). However, as Act-CHAMP reasons about the applied actions as well, it performed much better (see Table~\ref{table:table1}).

%\begin{figure}[b]
%\centering
%    \subfloat[Microwave]{\includegraphics[height=0.32\linewidth, width=0.23\textwidth]{figures/microwave_manipulation.png}} ~~%
%    \subfloat[Drawer]{\includegraphics[height=0.32\linewidth, width=0.23\textwidth]{figures/drawer_manipulation.png}}%
%\caption{Plots showing belief space [blue] and actual trajectories [orange] for microwave and drawer manipulation tasks using learned models. Error bars represent belief uncertainty.}
%\label{fig:figure7}
%% \vspace{-5pt}
%\end{figure}

\begin{figure}[b]
	\centering
	\begin{subfigure}{0.42\linewidth}
		\centering \includegraphics[height=0.8\linewidth, width=\linewidth]{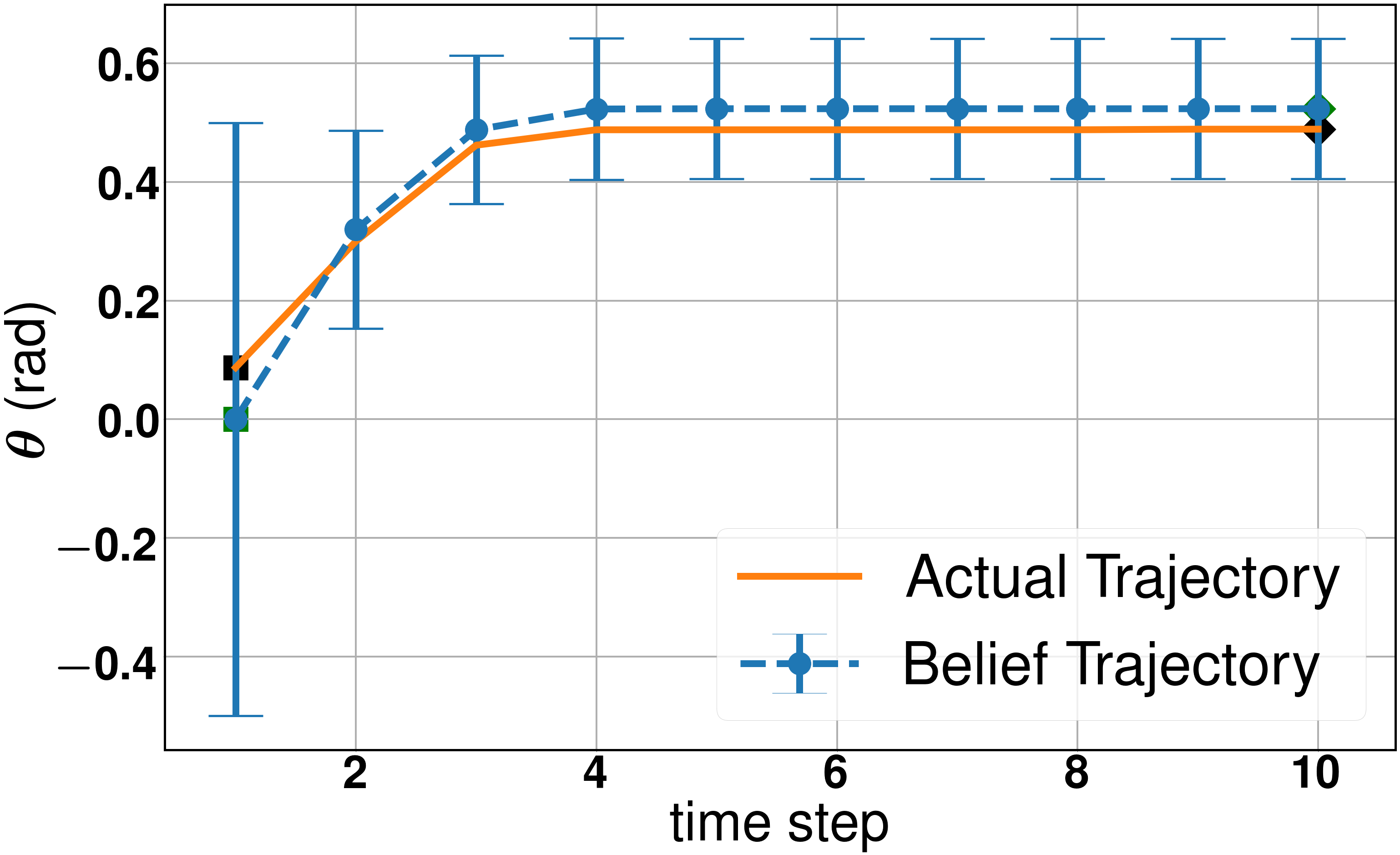}
		\caption{Microwave}
	\end{subfigure} \qquad%
	\begin{subfigure}{0.42\linewidth}
		\centering \includegraphics[height=0.8\linewidth, width=\linewidth]{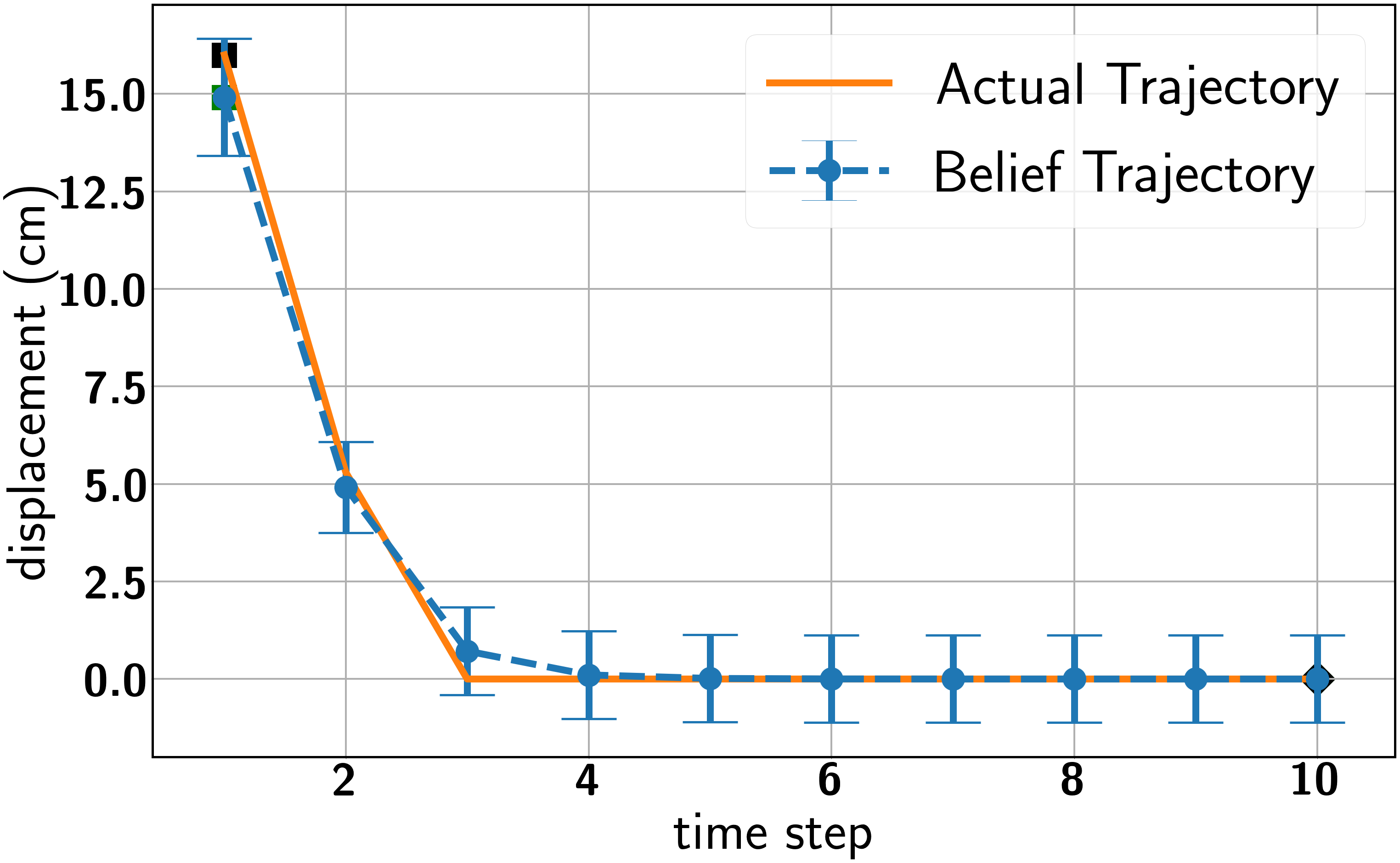}
		\caption{Drawer}
	\end{subfigure}%
	\caption{Plots showing belief space [blue] and actual trajectories [orange] for microwave and drawer manipulation tasks using learned models. Error bars represent belief uncertainty.}%    
	\label{fig:figure3}%
\end{figure}

\textbf{Without grasp:}
When actions are applied directly on the object (microwave door and the drawer, respectively), the majority of the applied actions are orthogonal to the axis of motion leading to low-information demonstrations.
% For the microwave, the observed significant relative displacement was found to be around $67.5\%$ sparser\footnotemark[3]\footnotetext[3]{$\%$ significant displacement = $\frac{\text{No. of time steps for which } \mb{y}_t \geq 0.005~\mathrm{units} }{\text{T (Total number of time steps in the time series)}} \times 100$} 
% in comparison to the microwave-with-grasp case (on average, only $27.71\%$of the observed displacements were found to be greater than the threshold of $0.005~\mathrm{rad}$ in comparison to $85.43\%$ for the microwave-with-grasp case). Extremely temporally sparse displacements combined with observational noise %$\Sigma_{y_{\mathrm{position}}} = 0.0055~m, \Sigma_{y_{\mathrm{orientation}}} = 1.4~\mathsf{rad}$ 
% results in poor model inference using either of the algorithms.
In such a case, while CHAMP almost completely failed to detect correct motion models for the microwave ($5\%$ success), Act-CHAMP was able to correctly detect models in almost one-third of the trials (see Table~\ref{table:table1}). 
% For drawer, the observed significant relative displacement was found to be around $58.92\%$ sparser than the drawer-with-grasp case ($19.05\%$ in comparison to $46.65\%$)). 
For the drawer, CHAMP falsely detected a changepoint and determined that the articulation motion model is composed of two separate prismatic articulation models with different model parameters (Figure~\ref{fig:figure2}). However, due to action-conditional inference, Act-CHAMP correctly classified the motion to be composed of only one articulation model (Figure~\ref{fig:figure2}, see Table~\ref{table:table1}).

% \begin{table}[t]
%     \centering
%     \begin{tabular}{ | c | c | c |}
%     \hline
%      \textbf{Experiments}& \thead{\textbf{CHAMP}} & \thead{\textbf{MICAH}} \\
%     \hline 
%     \textbf{Microwave w/ grasp} &  20/20 (100\%)~\footnotemark[4] & 20/20 (100\%)\\ 
%     \textbf{Microwave No-Actions} &  11/20 (55\%) & 12/20 (60\%)\\
%     %% WO grasp
%     \textbf{Microwave w/o grasp} & 1/20 (5\%) & 6/20 (30\%)\\%\hline
%     \hline
%     %% Drawer
%     \textbf{Drawer w/ grasp} & 20/20 (100\%) & 20/20 (100\%)\\ 
%     \textbf{Drawer No-Actions} & 4/20 (20\%) & 14/20 (70\%)\\ 
%     \textbf{Drawer w/o grasp} & 9/20 (45\%) & 15/20 (75\%)\\\hline
%     \end{tabular}
%     \caption{Comparison of CHAMP and MICAH across experiments}
%     \label{table:table1}
%     \vspace{-0.6cm}
% \end{table}
% \footnotetext[4]{$m/n$ represents $m$ number of correct model inferences out of $n$ experiments. We consider an inferred model to be correct if the number of detected changepoints matches the ground truth and the estimated local model parameters are within an $\epsilon$-radius ball of the ground truth parameters ($\epsilon = 0.05 \{ \mathrm{rad}/\mathrm{m} \}$)}

\begingroup
\setlength{\tabcolsep}{12pt} % Default value: 6pt
\begin{table*}[h]
    \centering
    \begin{tabular}{ |c | c | c | c | c | l |}
    \hline
     &&& \multicolumn{2}{c|}{\textbf{Changepoint Detection}} & \\ \cline{4-5}
     \textbf{Case}& \textbf{Object} & \textbf{Algorithms}& \thead{\textbf{Correct No.}} & \thead{\textbf{Position Error}} & \textbf{Error in Model Parameters}\\
    \hline 
    &&&&& \textbf{Center: } $0.046 \pm 0.024 ~\mathrm{m}$\\
    &\textbf{Microwave} &CHAMP & 20/20 (100\%) & $ 0.001 \pm 0.001$&\textbf{Axis:~ } $0.02 \pm 0.01~\mathrm{rad}$\\
    \textbf{With}&&&&&\textbf{Radius: } $0.027\pm0.024 ~\mathrm{m}$\\
    \cline{3-6}
    \textbf{grasp}&&&&& \textbf{Center: } $0.066 \pm 0.034~\mathrm{m}$ \\
    &&ActCHAMP & 20/20 (100\%)& $0.001 \pm 0.001$&\textbf{Axis: } $0.03\pm0.02~\mathrm{rad}$ \\
    &&&&&\textbf{Radius: }$0.013 \pm 0.059~\mathrm{m}$ \\
   \cline{2-6}
    &\textbf{Drawer} & CHAMP & 20/20 (100\%) & --- & \textbf{Axis: } $ 0.04 \pm 0.01~\mathrm{rad}$ \\ %\hline
    \cline{3-6}
    && ActCHAMP & 20/20 (100\%) & --- & \textbf{Axis:} $ 0.04 \pm 0.01~\mathrm{rad}$\\    
    
    %% No-op
    \hline
    &&&&& \textbf{Center: } $0.015 \pm 0.006~\mathrm{m}$ \\
    &\textbf{Microwave}&CHAMP & 11/20 (55\%) & $ 0.001 \pm 0.001$ &\textbf{Axis: } $0.01 \pm 0.01~\mathrm{rad}$\\
    \textbf{No}&&&&&\textbf{Radius: } $0.011 \pm 0.006~\mathrm{m}$\\
    \cline{3-6}
    \textbf{Actions}&&&&& \textbf{Center:} $0.019 \pm 0.009 ~\mathrm{m}$\\
    &&ActCHAMP & \textbf{14/20 (70\%)}& $ 0.001 \pm 0.001$& \textbf{Axis:} $0.01 \pm 0.01~ \mathrm{rad}$ \\
    &&&&& \textbf{Radius: } $0.002 \pm 0.014 ~\mathrm{m}$\\ 
    \cline{2-6}
    &\textbf{Drawer} & CHAMP & 4/20 (20\%) & --- & \textbf{Axis:} $0.03 \pm 0.01 ~\mathrm{rad}$\\
    \cline{3-6}
    &&ActCHAMP & \textbf{12/20 (60\%)}& --- & \textbf{Axis:} $0.03 \pm 0.01~\mathrm{rad}$\\
    
    %% WO grasp
    \hline
    &&&&& \textbf{Center: } $0.51 ~\mathrm{m}$ \\
    &\textbf{Microwave}&CHAMP & 1/20 (5\%) & $0.001$&\textbf{Axis: } $0.42~\mathrm{rad}$\\
    \textbf{Without}&&&&&\textbf{Radius: } $0.056~\mathrm{m}$\\
    \cline{3-6}
    \textbf{grasp}&&&&& \textbf{Center: } $0.328 \pm  0.125 ~\mathrm{m}$\\
    &&ActCHAMP & \textbf{6/20 (30\%)}& $0.001 \pm 0.001$& \textbf{Axis:} $0.58 \pm 0.20~ \mathrm{rad}$ \\
    &&&&& \textbf{Radius: } $0.164 \pm 0.103 ~\mathrm{m}$\\ 
    \cline{2-6}
    &\textbf{Drawer} & CHAMP & 9/20 (45\%) & --- & \textbf{Axis:} $0.18 \pm 0.01 ~\mathrm{rad}$\\
    \cline{3-6}
    &&ActCHAMP & \textbf{15/20 (75\%)}& --- & \textbf{Axis:} $0.14 \pm 0.04~\mathrm{rad}$\\
    % &&&&\\
    \hline
    \end{tabular}
    \caption{Model detection comparison}
    \label{table:table1}
%  \vspace{-10pt}
\end{table*}
\endgroup

%\begin{figure}[b]
%\centering
%    \subfloat[Stapler: $\{x, y, z\}$]{\includegraphics[height=0.35\linewidth, width=0.45
%    \linewidth]{figures/stapler_xyx_plot_new.png}} ~~~~~~~
%    \subfloat[Stapler: $\theta$ vs time]{\includegraphics[height=0.34\linewidth, width=0.45
%    \linewidth]{figures/stapler_theta_plot.png}}
%\caption{Planned trajectories for the stapler placement experiment. (Left) in $\{x, y, z\}$ (Right) Relative angle of the stapler arms over time.}
%\label{fig:figure8}
%%  \vspace{-5pt}
%\end{figure}

\begin{figure}[b]
	\centering
	\begin{subfigure}{0.42\linewidth}
		\centering \includegraphics[height=0.85\linewidth, width=\linewidth]{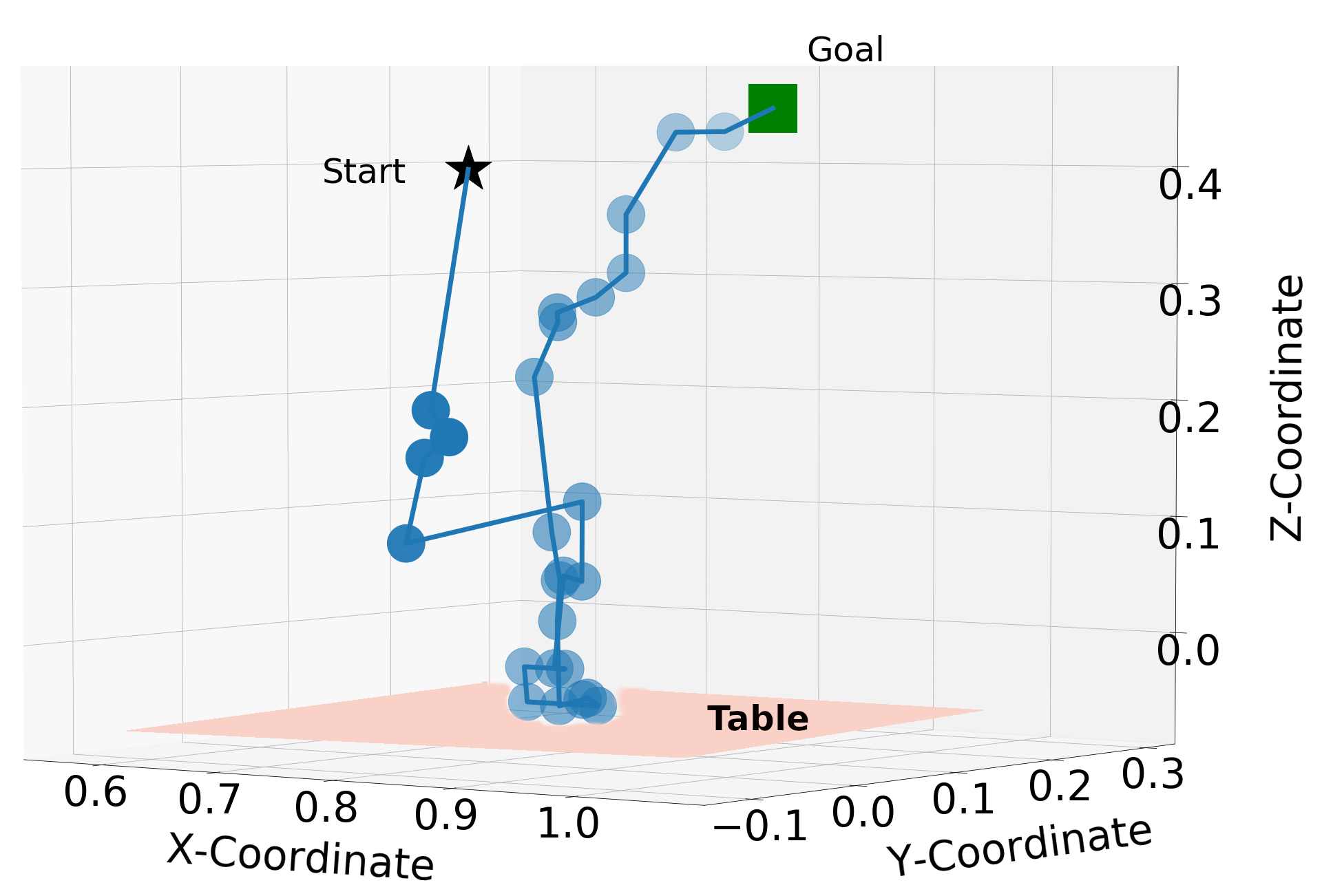}
			\caption{Stapler: $\{x, y, z\}$}
		\end{subfigure} \qquad
		\begin{subfigure}{0.42\linewidth}
			\centering \includegraphics[height=0.85\linewidth, width=\linewidth]{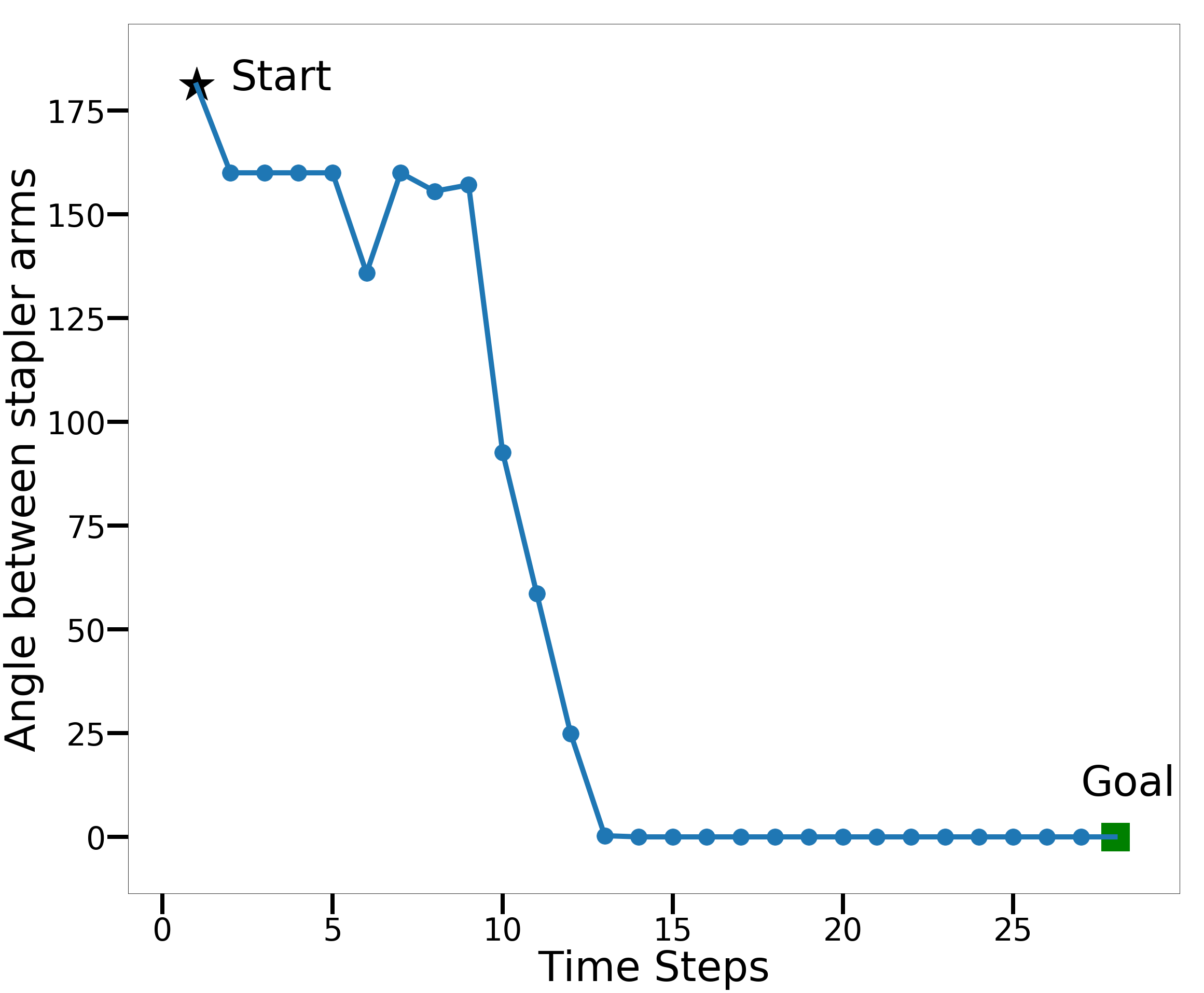}
			\caption{Stapler: $\theta$ vs time}
		\end{subfigure}
		\caption{Planned trajectories for the stapler placement experiment. (Left) in $\{x, y, z\}$ (Right) Relative angle of the stapler arms over time.}%    
		\label{fig:stapler_plan}%
	\end{figure}

\begin{figure*}[h]
    \centering
    \includegraphics[width=\linewidth, height=0.18\linewidth]{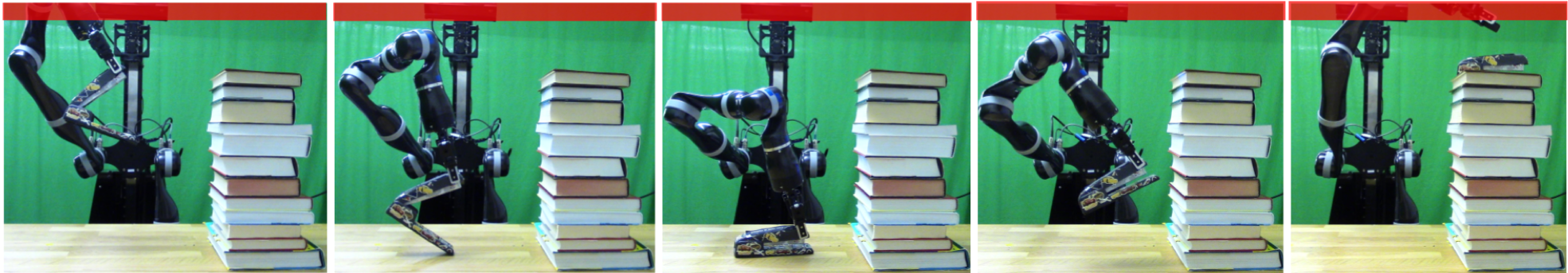}
    \caption{Snapshots showing the executed trajectory for the stapler placement task. The red region denotes the unreachable workspace for the robot's end-effector.}
    \label{fig:stapler_execution}
    % \vspace{-5pt}
\end{figure*}

\subsection{Object Manipulation Using Learned Models}
To test the effectiveness of the learned hybrid automata using MICAH, we used them to perform the tasks of opening and closing a microwave door and a drawer using a robot manipulator. We use the POMDP-HD planner \cite{jain2018efficient} to develop manipulation plans.
% We choose a POMDP planner as it can account for the state uncertainty due to modeling errors and perceptual errors while developing motion plans. 
Figure~\ref{fig:figure3} shows the belief space and actual trajectories for the microwave and drawer manipulation tasks. For both the objects, low final errors were reported: $0.05 \pm 0.01 ~\mathrm{rad}$ for the microwave and $0.005 \pm 0.003 ~\mathrm{m}$ for the drawer (average of 5 different tasks), validating the effectiveness of the learned automata.

\subsection{Leveraging Learned Models for Novel Manipulations}
Finally, we show that our learned models and planner are rich enough to be used to complete novel tasks under uncertainty that require intelligent use of object kinematics. To do so, we combine MICAH with the POMDP-HD planner for performing a manipulation task of placing a desk stapler at a target point on top of a tall stack of books. Due to the height of the stack, it is challenging to plan a collision-free path to deliver the stapler to the target location through a narrow corridor in the free configuration space of the robot; if the robot attempts to place the stapler at the target point while its governing kinematic model is revolute, the lower arm of the stapler will swing freely and collide with the obstacle. However, a feasible collision-free motion plan can be obtained if the robot first closes and locks the stapler (i.e. rigid articulation), and then proceeds towards the goal. To change the state of the stapler from revolute to rigid, the robot can plan to make contact with the table surface to press down and lock the stapler in a non-prehensile fashion.

As the task involves making and breaking contacts with the environment, we need to extend the learned hybrid motion model of the stapler to include local models due to contacts. We approximately define the contact state between the stapler and the table as to be either a line contact (an edge of the lower arm of the stapler in contact with the table), a surface contact (the lower arm lying flat on the table) or no contact. The set of possible local models for the hybrid task kinematics can be obtained by taking a Cartesian product of the set of possible discrete states for the stapler's hybrid automaton and the set of possible contact states between the stapler and the table. However, if the stapler is in the rigid mode, its motion would be the same under all contact states. Hence, a compact task kinematics model would consist of four local models---the stapler in revolute mode with no contact with the table, the stapler in revolute mode with a line contact with the table, the stapler in revolute mode with a surface contact with the table, and the stapler in rigid mode.
% Task dynamics can be approximated using a hybrid model consisting of four local models representing different system configurations --- stapler in revolute mode with no contact with the table, stapler in revolute mode with a line contact with the table, stapler in revolute mode with a surface contact with the table (lower arm lying flat on the table), and stapler in rigid mode (once in rigid articulation state, stapler's motion will be similar in all contact states). 
% In this task, we cannot directly actuate the stapler joint.
% Due to underactuation, finding cost-optimized paths using the direct transcription method for trajectory optimization, as proposed in the original POMDP-HD planner \cite{jain2018efficient}, is difficult. To ease out finding feasible paths in the task configuration space, we use the LQR-RRT* motion planner \cite{perez2012lqr} as the low-level motion planner in the hierarchical POMDP-HD planner, instead. 

Given a human demonstration of robot's interaction with the stapler as input, MICAH first learns a hybrid automaton for the stapler and then extends it to the hybrid task model using the provided task-specific parameters. Next, the POMDP-HD planner uses the learned task model to develop motion plans to complete the task with minimum final state uncertainty. Note that only the final Cartesian position for the stapler was specified as the target for the task and not the articulation state of the stapler (rigid/revolute). Motion plans generated by the planner are shown in Figure~\ref{fig:stapler_plan}. As can be seen from the plots, the planner plans to make contacts with the table to reduce the relative angle between the stapler arms and change the articulation model of the stapler. The plan drags the stapler along the surface of the table, indicating that it waits until it is highly confident that the stapler has become rigid before breaking contact. Making contacts with the table along the path also helps in funneling down the uncertainty in the stapler's location relative to the table in a direction parallel to the table plane normal, thereby increasing the probability of reaching the goal successfully. Figure~\ref{fig:stapler_execution} shows snapshots of the motion plan and actual execution of the robot performing the task.
% To compare with a baseline, we tried to use the LQR-RRT* planner \cite{perez2012lqr} for finding a path to the goal, but the planner failed to find a feasible path within the timeout limit.

%%%%%%%%%%%%%%%%%%%%%%%%%%%%%%%%%%%%%%%%%%%%%%%%%%%%%%%%%%%%%%%%%%%%%%%%%%%%%%%%
\section{Conclusion}
Robots working in human environments require a fast and data-efficient way to learn motion models of objects around them to interact with them dexterously. We present a novel method MICAH, that performs action-conditional model inference from unsegmented human demonstrations via a novel algorithm, Act-CHAMP, and then uses the resulting models to construct hybrid automata for articulated objects. Action-conditional inference enables articulation motion models to be learned with higher accuracy than the prior methods in the presence of noise and leads to the development of models that can be used directly for manipulation planning. Furthermore, we demonstrate that the learned models are rich enough to be used for performing novel tasks with such objects in a manner that has not been previously observed.
One advantage of using an action-conditional model inference approach over observation-only approaches is that it can enable robots to take informative exploratory actions for learning object motion models autonomously. Hence, future work may include the development of an active learning framework that can be used by a robot to autonomously learn the motion models of objects in a small number of trials. Another promising extension to the method can be to extend it to learn hybrid automata for articulated objects with multiple joints autonomously through noisy human demonstrations.

%Another promising direction is to infer more complex quasi-static articulation models between two objects in contact. 
%For example, a cup placed on a table cannot move towards the table plane as long as contact exists between them. This restriction can be viewed as though the contact has caused the two objects to be in a quasi-static articulation relationship. Taken together, these advances will move the state-of-the-art closer to a task-generic understanding of planning with complex objects in contact-rich settings.
%%%%%%%%%%%%%%%%%%%%%%%%%%%%%%%%%%%%%%%%%%%%%%%%%%%%%%%%%%%%%%%%%%%%%%%%%%%%%%%%
\section*{Acknowledgement}
This work has taken place in the Personal Autonomous Robotics Lab (PeARL) at The University of Texas at Austin. PeARL research is supported in part by the NSF (IIS-1724157, IIS-1638107, IIS-1617639, IIS-1749204) and ONR(N00014-18-2243).

%%%%%%%%%%%%%%%%%%%%%%%%%%%%%%%%%%%%%%%%%%%%%%%%%%%%%%%%%%%%%%%%%%
\bibliographystyle{IEEEtran}
\bibliography{IEEEabrv,paper}

\begin{thebibliography}{10}
\providecommand{\url}[1]{#1}
\csname url@rmstyle\endcsname
\providecommand{\newblock}{\relax}
\providecommand{\bibinfo}[2]{#2}
\providecommand\BIBentrySTDinterwordspacing{\spaceskip=0pt\relax}
\providecommand\BIBentryALTinterwordstretchfactor{4}
\providecommand\BIBentryALTinterwordspacing{\spaceskip=\fontdimen2\font plus
\BIBentryALTinterwordstretchfactor\fontdimen3\font minus
  \fontdimen4\font\relax}
\providecommand\BIBforeignlanguage[2]{{%
\expandafter\ifx\csname l@#1\endcsname\relax
\typeout{** WARNING: IEEEtran.bst: No hyphenation pattern has been}%
\typeout{** loaded for the language `#1'. Using the pattern for}%
\typeout{** the default language instead.}%
\else
\language=\csname l@#1\endcsname
\fi
#2}}

\bibitem{lygeros2012hybrid}
J.~Lygeros, S.~Sastry, and C.~Tomlin, ``Hybrid systems: Foundations, advanced
  topics and applications,'' \emph{under copyright to be published by Springer
  Verlag}, 2012.

\bibitem{pmlr-v100-gupta20a}
A.~Gupta, V.~Kumar, C.~Lynch, S.~Levine, and K.~Hausman, ``Relay policy
  learning: Solving long-horizon tasks via imitation and reinforcement
  learning,'' in \emph{Proceedings of the Conference on Robot Learning}, ser.
  Proceedings of Machine Learning Research, vol. 100.\hskip 1em plus 0.5em
  minus 0.4em\relax PMLR, 2020, pp. 1025--1037.

\bibitem{nagabandi2018neural}
A.~Nagabandi, G.~Kahn, R.~S. Fearing, and S.~Levine, ``Neural network dynamics
  for model-based deep reinforcement learning with model-free fine-tuning,'' in
  \emph{2018 IEEE International Conference on Robotics and Automation
  (ICRA)}.\hskip 1em plus 0.5em minus 0.4em\relax IEEE, 2018, pp. 7559--7566.

\bibitem{xu2020accelerating}
D.~Xu, M.~Agarwal, F.~Fekri, and R.~Sivakumar, ``Accelerating reinforcement
  learning agent with eeg-based implicit human feedback,'' \emph{arXiv preprint
  arXiv:2006.16498}, 2020.

\bibitem{kroemer2019review}
O.~Kroemer, S.~Niekum, and G.~Konidaris, ``A review of robot learning for
  manipulation: Challenges, representations, and algorithms,'' \emph{arXiv
  preprint arXiv:1907.03146}, 2019.

\bibitem{jain2018efficient}
A.~Jain and S.~Niekum, ``Efficient hierarchical robot motion planning under
  uncertainty and hybrid dynamics,'' in \emph{Conference on Robot Learning},
  2018, pp. 757--766.

\bibitem{Brunskill2008}
E.~Brunskill, L.~Kaelbling, T.~Lozano-Perez, and N.~Roy, ``{Continuous-State
  POMDPs with Hybrid Dynamics},'' \emph{Symposium on Artificial Intelligence
  and Mathematics}, pp. 13--18, 2008.

\bibitem{toussaint2008hierarchical}
M.~Toussaint, L.~Charlin, and P.~Poupart, ``Hierarchical pomdp controller
  optimization by likelihood maximization.'' in \emph{UAI}, vol.~24, 2008, pp.
  562--570.

\bibitem{niekum2015online}
S.~Niekum, S.~Osentoski, C.~G. Atkeson, and A.~G. Barto, ``Online bayesian
  changepoint detection for articulated motion models,'' in \emph{2015 IEEE
  International Conference on Robotics and Automation (ICRA)}.\hskip 1em plus
  0.5em minus 0.4em\relax IEEE, 2015, pp. 1468--1475.

\bibitem{pillai2015learning}
S.~Pillai, M.~R. Walter, and S.~Teller, ``Learning articulated motions from
  visual demonstration,'' \emph{arXiv preprint arXiv:1502.01659}, 2015.

\bibitem{martin2019coupled}
R.~Mart{\'\i}n-Mart{\'\i}n and O.~Brock, ``Coupled recursive estimation for
  online interactive perception of articulated objects,'' \emph{The
  International Journal of Robotics Research}, p. 0278364919848850, 2019.

\bibitem{sturm2011probabilistic}
J.~Sturm, C.~Stachniss, and W.~Burgard, ``A probabilistic framework for
  learning kinematic models of articulated objects,'' \emph{Journal of
  Artificial Intelligence Research}, vol.~41, pp. 477--526, 2011.

\bibitem{barragan2014interactive}
P.~R. Barrag{\"a}n, L.~P. Kaelbling, and T.~Lozano-P{\'e}rez, ``Interactive
  bayesian identification of kinematic mechanisms,'' in \emph{2014 IEEE
  International Conference on Robotics and Automation (ICRA)}.\hskip 1em plus
  0.5em minus 0.4em\relax IEEE, 2014, pp. 2013--2020.

\bibitem{perez2017c}
C.~P{\'e}rez-D'Arpino and J.~A. Shah, ``C-learn: Learning geometric constraints
  from demonstrations for multi-step manipulation in shared autonomy,'' in
  \emph{Robotics and Automation (ICRA), 2017 IEEE International Conference
  on}.\hskip 1em plus 0.5em minus 0.4em\relax IEEE, 2017, pp. 4058--4065.

\bibitem{subramani2018recognizing}
G.~Subramani, M.~Gleicher, and M.~Zinn, ``Recognizing geometric constraints in
  human demonstrations using force and position signals,'' in \emph{IEEE
  International Conference on Robotics and Automation (ICRA)}, 2018.

\bibitem{hausman2015active}
K.~Hausman, S.~Niekum, S.~Osentoski, and G.~S. Sukhatme, ``Active artic lation
  model estimation through interactive perception,'' in \emph{Robotics and
  Automation (ICRA), 2015 IEEE International Conference on}.\hskip 1em plus
  0.5em minus 0.4em\relax IEEE, 2015, pp. 3305--3312.

\bibitem{katz2008manipulating}
D.~Katz and O.~Brock, ``Manipulating articulated objects with interactive
  perception,'' in \emph{2008 IEEE International Conference on Robotics and
  Automation}.\hskip 1em plus 0.5em minus 0.4em\relax IEEE, 2008, pp. 272--277.

\bibitem{katz2013interactive}
D.~Katz, M.~Kazemi, J.~A. Bagnell, and A.~Stentz, ``Interactive segmentation,
  tracking, and kinematic modeling of unknown 3d articulated objects,'' in
  \emph{2013 IEEE International Conference on Robotics and Automation}.\hskip
  1em plus 0.5em minus 0.4em\relax IEEE, 2013, pp. 5003--5010.

\bibitem{martin2014online}
R.~M. Martin and O.~Brock, ``Online interactive perception of articulated
  objects with multi-level recursive estimation based on task-specific
  priors,'' in \emph{2014 IEEE/RSJ International Conference on Intelligent
  Robots and Systems}.\hskip 1em plus 0.5em minus 0.4em\relax IEEE, 2014, pp.
  2494--2501.

\bibitem{abbatematteo2019learning}
B.~Abbatematteo, S.~Tellex, and G.~Konidaris, ``Learning to generalize
  kinematic models to novel objects,'' in \emph{Proceedings of the Third
  Conference on Robot Learning}, 2019.

\bibitem{li2020category}
X.~Li, H.~Wang, L.~Yi, L.~J. Guibas, A.~L. Abbott, and S.~Song,
  ``Category-level articulated object pose estimation,'' in \emph{Proceedings
  of the IEEE/CVF Conference on Computer Vision and Pattern Recognition}, 2020,
  pp. 3706--3715.

\bibitem{fearnhead2007line}
P.~Fearnhead and Z.~Liu, ``On-line inference for multiple changepoint
  problems,'' \emph{Journal of the Royal Statistical Society: Series B
  (Statistical Methodology)}, vol.~69, no.~4, pp. 589--605, 2007.

\bibitem{torr2000mlesac}
P.~H. Torr and A.~Zisserman, ``Mlesac: A new robust estimator with application
  to estimating image geometry,'' \emph{Computer vision and image
  understanding}, vol.~78, no.~1, pp. 138--156, 2000.

\bibitem{pauwels_simtrack_2015}
K.~Pauwels and D.~Kragic, ``Simtrack: A simulation-based framework for scalable
  real-time object pose detection and tracking,'' in \emph{IEEE/RSJ
  International Conference on Intelligent Robots and Systems}, Hamburg,
  Germany, 2015.

\end{thebibliography}

\end{document}